\documentclass{article}

\usepackage[final, nonatbib]{nips}

\usepackage[utf8]{inputenc} % allow utf-8 input
\usepackage[T1]{fontenc}    % use 8-bit T1 fonts
\usepackage{hyperref}       % hyperlinks
\usepackage{url}            % simple URL typesetting
\usepackage{booktabs}       % professional-quality tables
\usepackage{amsfonts}       % blackboard math symbols
\usepackage{amsmath}
\usepackage{nicefrac}       % compact symbols for 1/2, etc.
\usepackage{microtype}      % microtypography
\usepackage{multirow}
\usepackage{array}
\usepackage{graphicx}
\usepackage{color}
\usepackage{fancyhdr}
\DeclareMathOperator*{\argmax}{arg\,max}

\title{Rethinking Machine Learning Development and Deployment for Edge Devices}

\author{
  Liangzhen Lai \\
  Arm Inc. \\
  \texttt{liangzhen.lai@arm.com} \\
  \And
  Naveen Suda \\
  Arm Inc.\\
  \texttt{naveen.suda@arm.com} \\
}

\begin{document}
% \nipsfinalcopy is no longer used

\maketitle

\begin{abstract}
Machine learning (ML), especially deep learning is made possible by the availability
of big data, enormous compute power and, often overlooked, development tools
or frameworks.
As the algorithms become mature and efficient, more and more ML inference is 
moving out of datacenters/cloud and deployed on edge devices.
This model deployment process can be challenging as the deployment 
environment and requirements can be substantially different from those during 
model development. In this paper, we propose a new ML development and deployment 
approach that is specially designed and optimized for inference-only deployment
on edge devices. We build a prototype and demonstrate that this approach 
can address all the deployment challenges and result in more efficient and 
high-quality solutions.
\end{abstract}

\section{Introduction}
\label{sec:introduction}

Deep neural networks (DNNs) have demonstrated near-human accuracy in a wide range
of applications, including image classification, speech recognition and
natural language processing. The availability of large training datasets and
compute power has been key enablers for this breakthrough.
Arguably and often overlooked, the availability of ML development tools
or frameworks, such as Caffe~\cite{jia2014caffe} and
TensorFlow~\cite{abadi2016tensorflow}, is also important as they allow 
more developers to easily experiment with new networks/algorithms and quickly 
evaluate them for new types of applications.

There are a lot of benefits in running neural network inference locally at the
edge~\cite{mltoedge,tinyml}. A lot of research efforts have been focusing on designing more
efficient networks~\cite{iandola2016squeezenet,howard2017mobilenets,huang2017densely},
more compact numerical representation~\cite{lin2016fixed, chung2017accelerating,
lai2017deep, settle2018quantizing, binarynet} and more powerful inference engines
both in terms of dedicated hardware~\cite{jouppi2017datacenter,chen2016eyeriss} and optimized 
software~\cite{lai2018cmsis, chetlur2014cudnn}. Together, these make it possible
to run DNN-based solutions on edge and even deeply embedded devices~\cite{zhang2017hello}.

Despite the good progress on demonstrating the capabilities of running DNNs at 
the edge, the NN model deployment process remains as both a challenging and tedious process. 
In most scenarios, the deployment targets can be embedded devices with limited memory 
and compute resources. Running the NN inference inside a full-blown framework environment
may be impractical and unnecessary. Most existing solutions implement either an inference-only
runtime~\cite{tflite} or a static compilation flow~\cite{tfxla}. But there can still be some challenges
when using these solutions to deploy a pre-trained and pre-optimized NN model from ML 
frameworks. These challenges are mostly created by the difference between framework 
environment and inference environment, which includes: 
\begin{itemize}
\item Operator availability: ML frameworks are usually designed to be flexible and can include
many experimental or customized operators. This enables ML developers to experiment with new 
types of network architecture, but at the same time makes it difficult to deploy these models
with inference-only implementation, which is typically optimized for efficiency. 
\item Operator behavior: Inference-only implementations, especially hardware implementations
can be novel in their approaches, e.g., the use of Winograd convolution~\cite{lavin2016fast},
different numerical representation~\cite{chung2017accelerating, sharma2017bit}, data 
compression~\cite{han2015deep} or even non-digital 
computation~\cite{shafiee2016isaac, akopyan2015truenorth}.
The operators in these implementations are likely to be different from the standard 
32-bit floating point (fp32) based implementations inside most ML frameworks. 
\end{itemize}

%+ Tool infrastructure is important to enable developer to explore more applications.
In this work, we explore an alternative approach of ML development and deployment  
targeting deployment at the edge. We build a prototype targeting Arm Cortex-M CPUs
with CMSIS-NN~\cite{lai2018cmsis} and demonstrate that this approach can greatly 
simplify the model deployment process and generate solutions with better efficiency 
and quality.

\section{Current Approach}
\label{sec:curr_approach}

\begin{figure}
\centering
\includegraphics[width=0.9\textwidth]{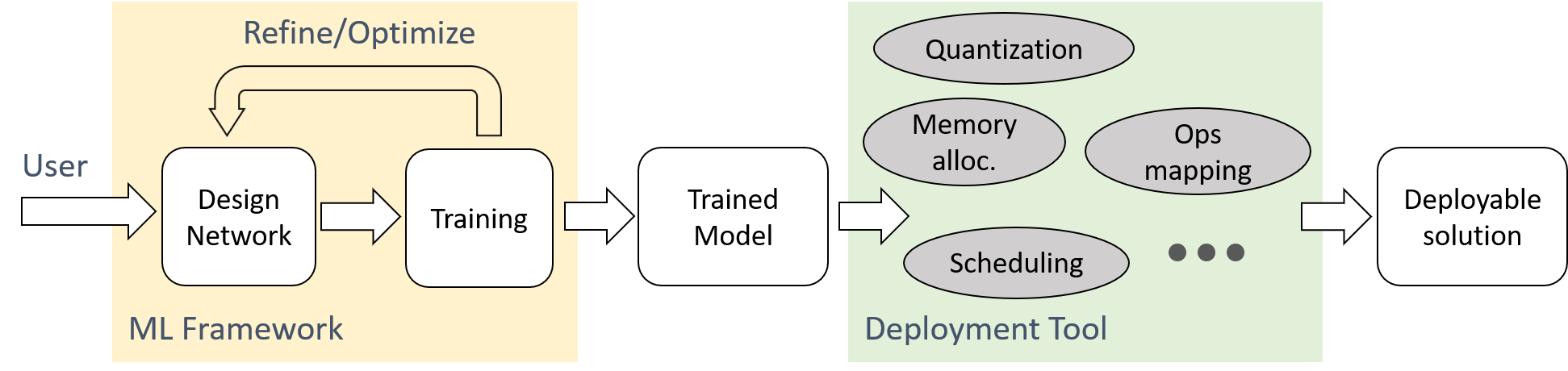}
\caption{\label{fig:currappr}Current ML development and deployment approach.
The shaded blocks are vendor-specific parts.}
\end{figure}

Current ML development and deployment approach is shown in Figure~\ref{fig:currappr}.
Application developers use the ML frameworks to construct their
NN models, perform training and evaluate the accuracy. Based on the evaluation results,
they can go back and refine or optimize their models. After this development stage,
a trained model is generated by the framework. The deployment tool, typically offered
by the deployment platform vendor, takes this trained model as the input and constructs
a deployable solution that runs on the target platform. 

This deployment process can be divided into two parts. The first part is the model 
execution, which includes parsing the NN model graph, mapping the operators to the 
available implementation, and then re-constructing the execution graph with proper resource 
allocation. The second part is the parameter conversion which takes the trained weights 
or parameters and converts them into the format and values that can be fed into their 
operator implementation while retaining or closely replicating the expected behavior.

The difference in operator availability and behavior, as discussed earlier, poses challenges 
to these two parts of the deployment process. Standard formats~\cite{ONNX} or APIs~\cite{nnapi} 
can be a solution to the operator availability issue. This solution, however, also forces the 
inference implementation to be generic (e.g., to support all of the standard operators) or with
enough flexibility rather than being able to optimize for specialization. 
Since the ML algorithms and frameworks may evolve over time, these standards can also be a moving 
target, which makes it difficult for the implementation, especially the hardware implementation, 
to be future-proof. 
The operator behavior issue is another limiting factor that restricts the implementation design
space. The implemented operators have to closely replicate the operators that are in the 
ML frameworks, which may not be the most implementation friendly and efficient options.

Conceptually, there are three ML models during the entire ML development and deployment process:
the model that is designed and optimized by the developer, the model that is trained in ML framework
and the model that is deployed on the device.
The design optimization of these three models using current approach can be represented 
by the following equation:
\begin{equation}
\label{equ:curr}
\argmax(model) = \argmax(train) \approx \argmax(deploy)
\end{equation}
$\argmax(model)$ represents the process of model optimization during ML development.
In the current approach, where the user specifies and constructs the NN model inside the 
framework using native operators, this model is the same as the model that gets trained by
the framework. Therefore, all the model optimization is equivalent to optimizing the
model that is trained in ML framework. This is represented by the first half of the equation, 
i.e., $\argmax(model) = \argmax(train)$. However, as discussed earlier, the deployed model 
generated by the deployment tools may not be exactly the same as the trained model.
As a result, the model optimization during the development process ($\argmax(model)$)
is only an approximation of optimizing the deployed model ($\argmax(deploy)$). This is
represented by the second half of the equation.

The other problem for this approach is the deployment uncertainty. The accuracy of the deployed
model is unknown at model development time, which makes it difficult for the developer to
control and guarantee solution quality. This accuracy uncertainty, a.k.a. accuracy loss,
may be quantified empirically with some benchmark models. But it is still difficult to 
guarantee whether this can be generalized to all kinds of models.

\section{Proposed Approach}
\label{sec:new_approach}

If the target is deployment on the edge devices, the deployment quality (i.e.,
$\argmax(deploy)$) should be the primary optimization target, rather than the 
model specified and trained in the ML framework (i.e., $\argmax(train)$).

\begin{figure}
\centering
\includegraphics[width=0.9\textwidth]{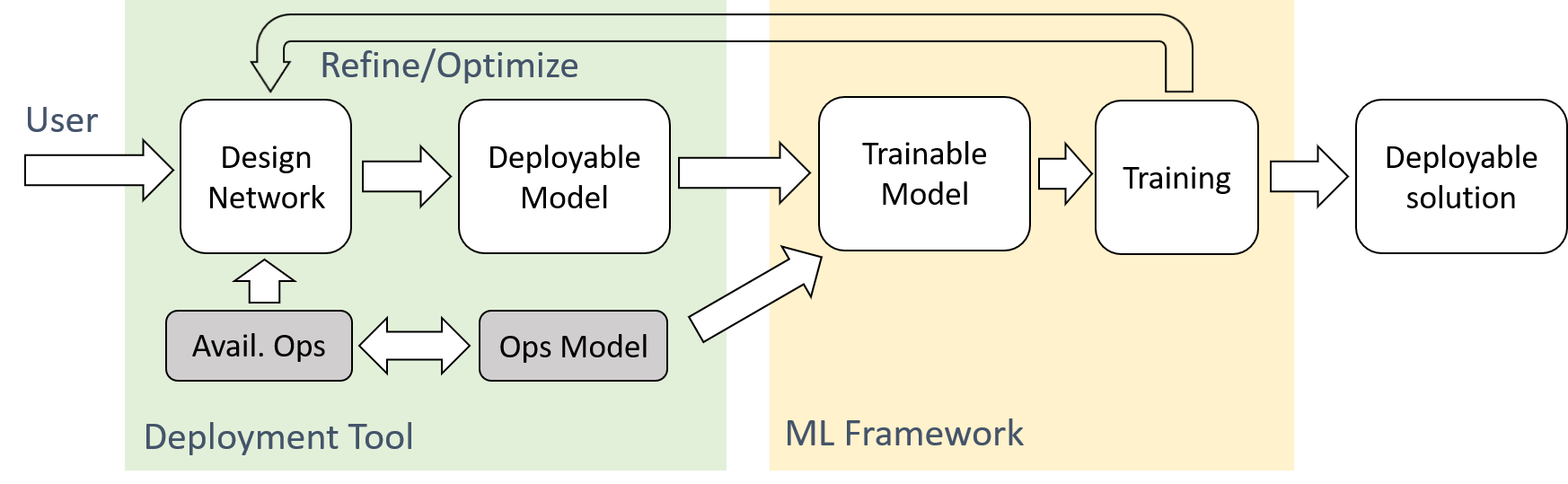}
\caption{\label{fig:ourappr}Proposed ML development and deployment approach.
The shaded blocks are vendor-specific parts.}
\end{figure}

To achieve this, we proposed the ML development and deployment approach shown in
Fig.~\ref{fig:ourappr}.
Unlike current approach, the users, which are the solution developers, will
specify their network models using the deployment tool with an operator library from
the deployment platform vendor. The deployment tool will generate a deployable
but untrained solution, i.e., a model that can run on their platform, but without 
the trained parameters. As part of the deployment tool, the vendor should also be
responsible to create an operator model library that can represent their operators 
and be used to create a trainable model inside the ML framework. 
This trainable model should have a forward inference path that behaves exactly the 
same as the implemented operators, and a backward path for the ML framework to 
perform training in order to generate the appropriate weights and parameters. 
In this approach, the validation results from the ML framework will be the same as 
the deployment results, so that the user can use these results to refine or 
further optimize their network models.

This proposed approach can be represented as the following equation:
\begin{equation}
\argmax(model) = \argmax(deploy) \approx \argmax(train)
\end{equation}
where the model constructed by the user is the same as the model that is going
to be deployed on the device, i.e., $\argmax(model) = \argmax(deploy)$.
In this approach, the training target is the trainable model generated by the 
deployment tool rather than the constructed or deployed model. Therefore, all
the training optimization is only an approximation of optimizing the actual
model, which is represented by the second half of the equation.

This approach addresses the operator availability issue by moving the network specification
process into the deployment environment. This can also give accurate estimates of the inference 
latency, memory footprints and energy cost, which helps network designer to pick the more 
efficient operators for the targeting platform~\cite{lai2018not}. 

The operator behavior issue in this approach becomes the requirements to make sure that the
implemented operators have trainable representations inside the ML framework.
This is a relatively easier problem than making sure that the implemented operators
behave similarly to the equivalent ones in the ML framework, as the solutions for the
latter problem is a subset of the former one.

The other potential advantage of this approach is that the trainable model inside the ML 
framework can also be used as the golden model for debug and validation purposes.
This avoids the need to implement an execution model of the deployment platform from scratch.
Re-using the trainable model can also leverage all the already well-optimized computation 
routines/kernels inside the framework.

\section{Proof-of-Concept Experiments}
\label{sec:results}

\subsection{Operator Model Implementation}

To validate the proposed approach, we implement a prototype of an operator library and 
an operator model library targeting Arm Cortex-M CPUs with CMSIS-NN~\cite{lai2018cmsis} and
TensorFlow~\cite{abadi2016tensorflow}. In this work, we refer this prototype as 
KANJI.

\begin{figure}[h]
\centering
\includegraphics[width=0.6\textwidth]{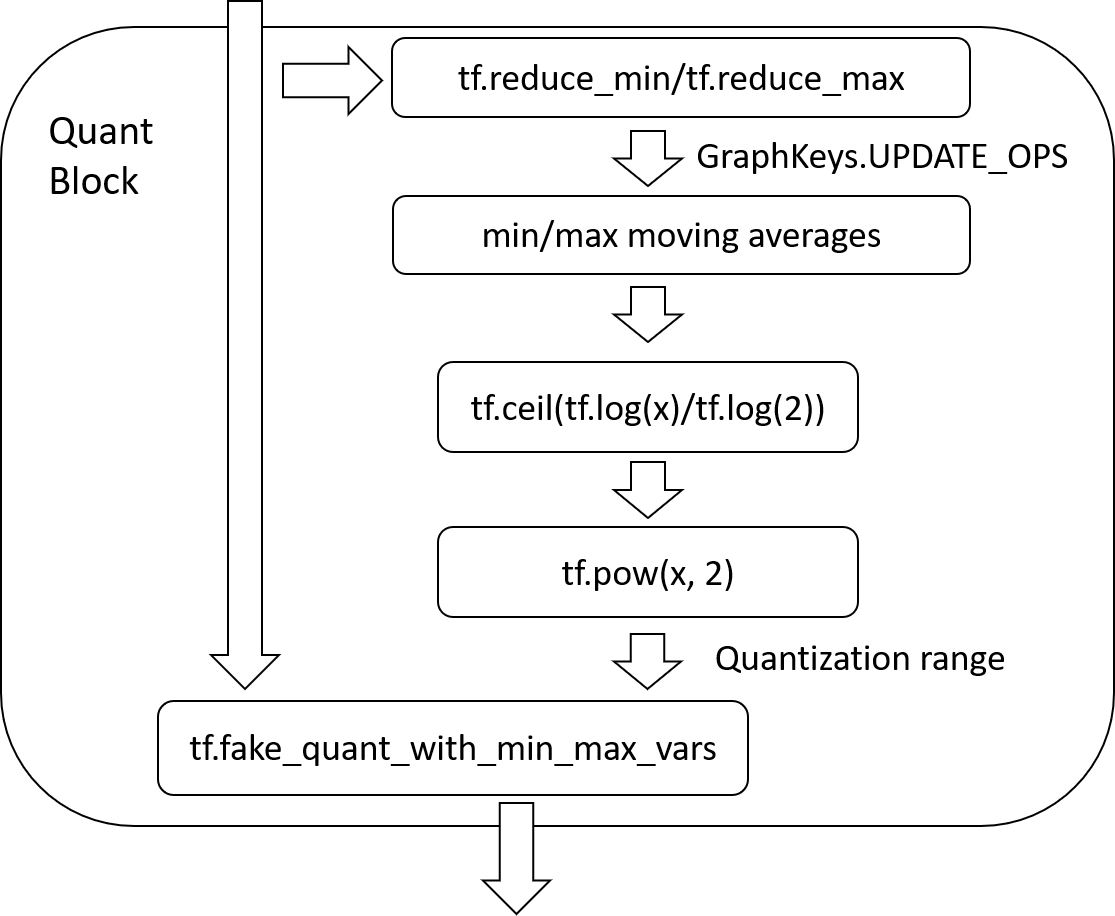}
\caption{\label{fig:quant}Convolution Model Implementation using TensorFlow operators
for CMSIS-NN}
\end{figure}

The computation kernels in CMSIS-NN implements int8 fixed-point computation. Therefore, 
the quantization is performed symmetrically around zero. The quantization step is forced
to be power-of-2 so that the transformation between different quantization formats becomes
simple bit-level shifting. 
To replicate the same quantization setup as that in CMSIS-NN, we design a specific quantization 
block as shown in Fig.~\ref{fig:quant}.
Inspired by how batch normalization is implemented, this quantization block keeps track of
the data distribution of the input and sets the quantization range accordingly.
The value quantization is performed by using the $FakeQuant$ operator from TensorFlow.

The operator library includes image pre-processing, convolution layer, fully-connected layer, max 
pooling and ReLU. Some of the layers, e.g., max pooling and ReLU, have the same behaviors in
TensorFlow and CMSIS-NN, so the default TensorFlow operators are used. 
For other layers, we implement the computation using the quantization block and built-in
TensorFlow operators. 

\begin{figure}[h]
\centering
\includegraphics[width=0.6\textwidth]{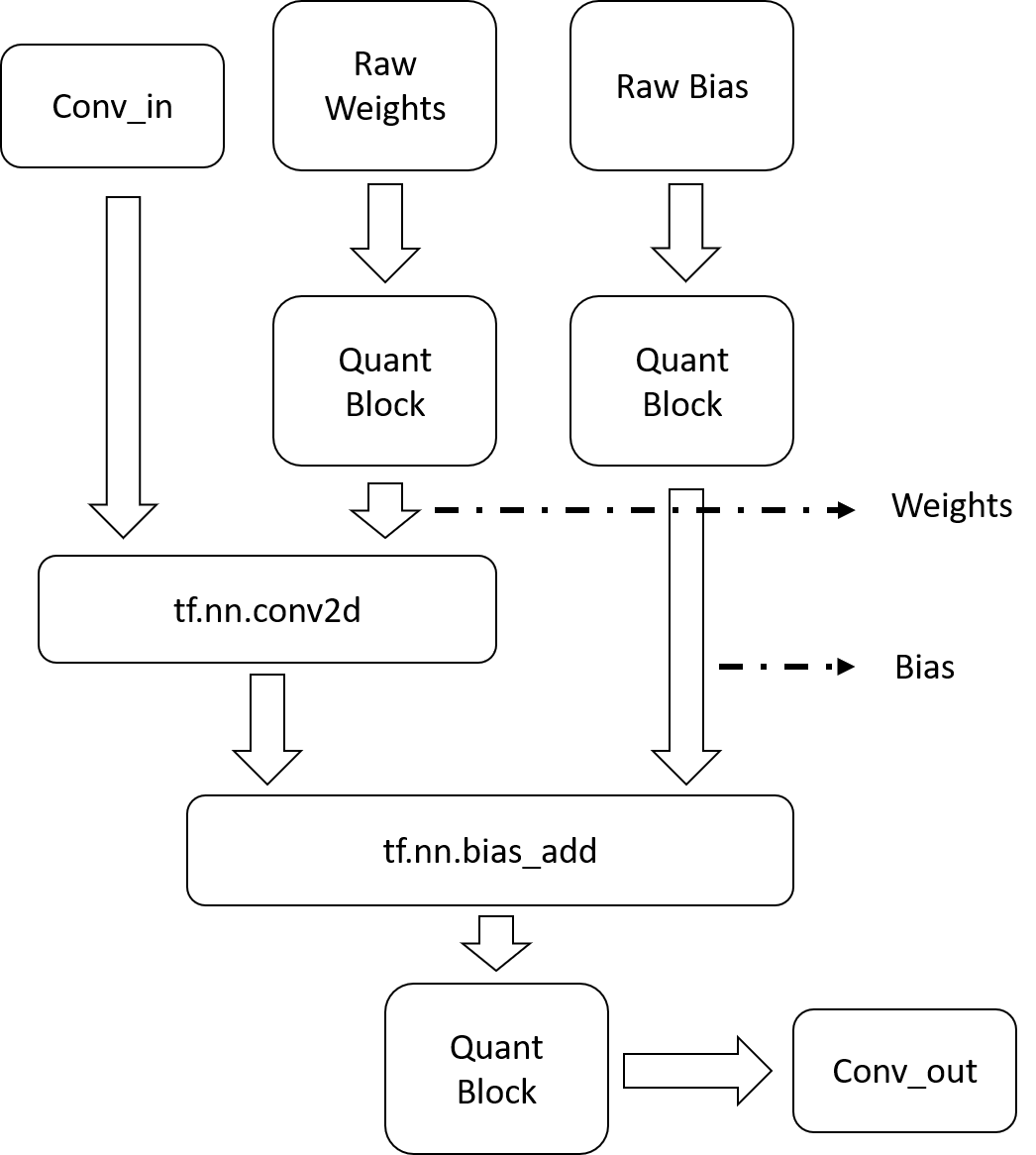}
\caption{\label{fig:conv}Convolution Model Implementation using TensorFlow Operators
for CMSIS-NN}
\end{figure}

An example of the convolution operator implementation is shown in Fig.~\ref{fig:conv}.
In this convolution model, both raw weights and bias are fp32 trainable parameters.
The quantization block is tracking the moving average of the data min/max and adjust
the quantization range accordingly.
The int8 weights and bias values used by CMSIS-NN convolution function will be the tensor 
value after the quantization block. 
The quantization ranges for inputs, weights, bias and outputs are used to determine 
the $out\_shift$ and $bias\_shift$ parameters for the convolution functions in CMSIS-NN.
We also validate that the operator behavior in TensorFlow is identical to the behavior
of the CMSIS-NN kernel running on Cortex-M CPUs.

\subsection{Experimental Results}

In this section, we will evaluate the implemented prototype KANJI. In particular,
we want to evaluate the impacts of the following:
\begin{itemize}
\item $\argmax(model) \approx \argmax(train)$ vs. $\argmax(model)=\argmax(train)$: 
comparing the capabilities (i.e., accuracy) of the same NN model trained with different 
approaches. This is described in Section~\ref{sec:acc}.
\item $\argmax(model)=\argmax(deploy)$ vs. $\argmax(model)\approx\argmax(deploy)$: 
studying the potential improvements on inference efficiency and quality using KANJI. 
This is described in Section~\ref{sec:op_dependence}.
\end{itemize}

\subsubsection{Accuracy Impacts}
\label{sec:acc}
To evaluate the accuracy, we implement the CNN example described in
CMSIS-NN~\cite{lai2018cmsis} using both default fp32 based training
and KANJI. The CNN example is designed for CIFAR-10 dataset and has
3 convolution layers, 3 pooling layers and 1 fully-connected layer. 
The numbers of output channels for the three convolution layers are 32, 32 
and 64, respectively. The data augmentation, learning rate control
and optimizer setup is the same as the CIFAR-10 example in TensorFlow.

The accuracy results are summarized in Table~\ref{tbl:cifar10}. We 
also repeat the experiments with different network sizes by changing 
the number of output channels in the convolution layers.
In most cases, the results are similar between the fp32 model and
the KANJI int8 model. We observe a trend that KANJI performs relatively 
better on smaller network sizes than larger ones. But this may be an artifact
of the training setup optimizer and learning rate setup, which will be discussed
in Section~\ref{sec:training_control}.
 
\begin{table}
\caption{\label{tbl:cifar10} Accuracy results for CIFAR-10 Dataset 
with different network sizes.}
\centering
\begin{tabular}{|c|c|c|c|}
\hline
Number of Conv Channels &  fp32  &  int8 KANJI \\ 
\hline
(16, 16, 32)            &  77.3\%  &  78.5\% \\
\hline
(32, 32, 64)            &  80.8\%  &  81.9\% \\
\hline
(48, 48, 96)            &  83.1\%  &  83.0\% \\
\hline
(64, 64, 128)           &  84.2\%  &  83.7\% \\
\hline
\end{tabular}
\end{table}

We also carry out the experiments with larger networks and datasets.
We use VGG network architecture and perform the training for both
CIFAR-100 and tiny-imagenet datasets. The accuracy results are 
summarized in Table~\ref{tbl:other}. 
Similar to CIFAR-10 results, the accuracy results are very similar
for int8 KANJI models and fp32 models.

\begin{table}
\caption{\label{tbl:other} Accuracy results with VGG for CIFAR-100 
and tiny-imagenet dataset.}
\centering
\begin{tabular}{|c|c|c|}
\hline
Data set           & fp32 & int8 KANJI \\
\hline
CIFAR-100 top1     & 59.5\% & 59.6\%       \\  
\hline
CIFAR-100 top3     & 77.9\% & 77.4\%       \\
\hline
tiny-imagenet top1 & 42.0\% & 42.0\%       \\
\hline
tiny-imagenet top3 & 60.0\% & 60.1\%       \\
\hline
\end{tabular}
\end{table}

\subsubsection{Input-dependent vs. Input-independent Operators}
\label{sec:op_dependence}
One important advantage of the proposed approach is that it allow users to 
use the operators that can be implemented more efficiently, rather than the default 
operators in the ML frameworks. In this section, we show through examples that
our proposed approach can generate more efficient inference solutions with similar or 
better accuracy.

An example is the use of input-independent operators instead of the input-dependent ones.
For example, it is known that normalizing the input data can improve the 
effectiveness of the training and the model accuracy.
One of the most popular image pre-processing operators in TensorFlow is  
$per\_image\_standardization$ where each image channel is shifted and scaled to have
zero mean and unit variance. This is an input-dependent operator as the amount of
shifting and scaling depends on the input data distribution. These values can vary
for different input images and have to be calculated for each inference run. 
Caffe extracts the mean value for each pixel by scanning the entire training set 
and storing them in the $mean.binaryproto$ file and included as part of the model
parameters. At inference time, these mean values are subtracted from the input image.
This image pre-processing operator is input independent as the same mean pixel 
value is subtracted regardless of what the input image is.  
In KANJI, we implement an operator that is similar to $batch\_norm$ but
without offset (i.e., $\beta$) and scale (i.e., $\gamma$), so the data normalization
process is input-independent and the storage overhead is kept minimal.
We also force the shifted mean (i.e., $\mu$) to be integer, and scaling
(i.e., $\sigma$) to be power-of-2, so that the process becomes simple integer subtraction
and shifting. The details of different image pre-processing options are listed
in Table~\ref{tbl:norm}

\begin{table}
\caption{\label{tbl:norm} Different image pre-processing operators.}
\centering
\begin{tabular}{|c|c|c|c|}
\hline
Image Pre-Processing Operator &  Input Dependence & Memory Overhead & Runtime Overhead \\
\hline
$per\_image\_standardization$ &  Yes              &    Low          &  High            \\
\hline
$mean.binaryproto$            &  No               &    High         &  Low             \\
\hline
$batch\_norm$-like            &  No               &    Low          &  Low             \\
\hline
\end{tabular}
\end{table}

The other example is asymmetric quantization vs. symmetric quantization. Quantization,
in this context, refers to the process of mapping a floating-point value to an integer
value. The relationship between the floating-point value ($value\_fp$) and quantized
integer value ($value\_int$) can be represented with the quantization range
($[min\_fp, max\_fp)$) and quantization step ($step\_fp$) by the following equation:
\begin{equation}
value\_fp = min\_fp + (value\_int - min\_int) * step\_fp
\end{equation}

There are different ways to perform quantization. The simplest way is symmetric quantization
with power-of-2 step value, i.e., $min\_fp + max\_fp = 0$ and $step\_fp = 2^n$ where
n is a integer value. This is also referred as fixed-point quantization and is used by CMSIS-NN.
The other popular way is asymmetric quantization, e.g., supported by Android NNAPI~\cite{nnapi} 
where quantization range and step size can be arbitrary value.

The symmetry of the quantization scheme affects the quantized computation. 
For example, quantized matrix multiplication can be implemented as regular integer matrix 
multiplication if the quantization is symmetric. If the quantization is asymmetric, implementation 
similar to GEMMLOWP~\cite{gemmlowp} can be used, where the bulk part of the computation is still 
regular integer matrix multiplication, and additional routines are needed to compute the impacts of
non-zero offsets. To quantify the overhead of these additional routines, we also implement 
GEMMLOWP-like matrix multiplication routine for asymmetric quantization. Experiments on Cortex-M 
CPUs show that the runtime overhead is about 15\% for typical network sizes.

The other important part is how to quantize the computation outputs (i.e., accumulators).
One way to quantize the outputs is to find out the min and max values and apply the quantization 
accordingly. This is an input-dependent process and usually requires to store all raw outputs 
temporarily. This can result in 2-4X increase in runtime memory footprint. On the contrary, CMSIS-NN 
forces the quantization step to be power-of-2 so that the output quantization can be done on-the-fly 
with simple shifting.

These implementation friendly operators can improve the inference efficiency. They can also affect the
model accuracy. We repeat the CIFAR-10 experiments with different image pre-processing techniques
and quantization schemes. The accuracy results are shown in Table~\ref{tbl:quant}. Among the three
image pre-processing operators, the $batch\_norm$-like operator gives the best accuracy.
The quantization does not seem to degrade the accuracy. In the case of KANJI, where quantization is 
accounted for during training, the accuracy may even be higher than the fp32 counterpart.

\begin{table}
\caption{\label{tbl:quant} Network accuracy on CIFAR-10 with different image pre-processing and 
quantization.}
\centering
\begin{tabular}{|c|c|c|}
\hline
Input Pre-Processing           & Quantization & Accuracy \\
\hline
$per\_image\_standardization$ & None (fp32) & 81.2\% \\
\hline
$per\_image\_standardization$ & 8-bit asymmetric & 81.2\% \\
\hline
$mean.binaryproto$            & None (fp32) & 81.5\% \\
\hline
$mean.binaryproto$            & 8-bit, symmetric & 81.5\% \\
\hline
$batch\_norm$-like            & None (fp32) & 81.8\% \\
\hline
$batch\_norm$-like            & KANJI & 81.9\% \\
\hline
\end{tabular}
\end{table}

\section{Discussion}
\label{sec:discussion}

\subsection{Model Training}
\label{sec:training_control}
\begin{figure}[h]
\centering
\includegraphics[width=0.95\textwidth]{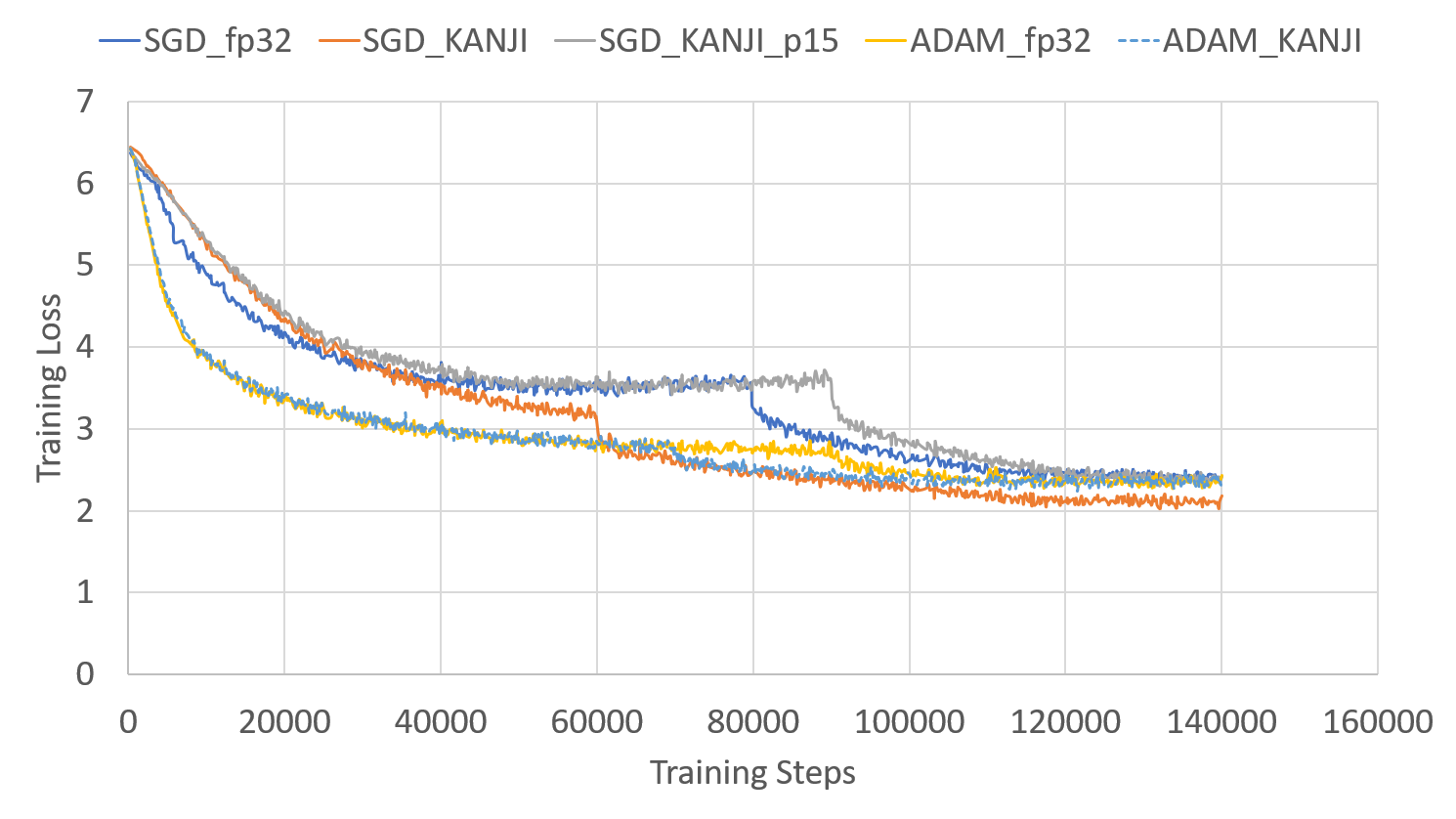}
\caption{\label{fig:loss}Training loss for tiny-imagenet with different
optimizer and initial learning rate.}
\end{figure}

The training process using the proposed approach can be different from training a 
native fp32 model. This difference can include the effectiveness of different training
setups, such as loss functions, optimizers, batch sizes, and learning rates.
For example, the training loss curves of native fp32 training and KANJI using
different optimizers and initial learning rates are plotted in
Fig.~\ref{fig:loss}. For SGD optimizer, loss reduces faster but saturates earlier
for fp32 training (SGD\_fp32), comparing to KANJI (SGD\_KANJI), 
even though they have the same initial learning rate of $0.1$. 
Since the training target in KANJI is an 8-bit model, there may be some damping effects 
in gradient approximation and updates, making the effective learning rate smaller. 
We repeat the experiments for KANJI with an increased initial learning rate of $0.15$
(SGD\_KANJI\_p15). In this case, the loss saturates to a similar level as SGD\_fp32.
This difference in learning rate is not observed with ADAM optimizer, with 
the loss curves for both fp32 training (ADAM\_fp32) and KANJI (ADAM\_KANJI) 
following similar trends. 

The configuration and hyperparameter setting during the training can have
big impact on the final model accuracy. 
In our experiments, we try to keep the same setup for KANJI and native fp32
training to get fair comparisons. Comparing to the training process using
the proposed approach, the training process of native fp32 model is relatively 
better understood and optimized.
More learning and research of the training process could further improve the 
solution quality of the proposed approach.

The other noticeable difference is the training time. Training using the proposed
approach is likely to be slower. 
Based on our experiments, training time per step in KANJI is about 20\% to 25\% slower
than training the fp32 model. But the total training time comparison may be
different and will also depend on how the model converges. If the training efficiency
is critical, a hybrid approach can also be used where the model is pre-trained 
in fp32 and fine-tuned using the proposed approach.  

\subsection{Pre-trained Models}
\label{sec:pretrained}

Although the proposed approach shown in Fig.~\ref{fig:ourappr} assumes that the 
entire process is starting from scratch, in many scenarios, it is both necessary 
and beneficial to be able to incorporate a pre-trained model into the development 
and deployment process. The hybrid training approach discussed in previous section 
is one example.

There are also many well-trained and well-optimized NN models that can be directly
used or re-targeted (e.g., through transfer learning).
Being able to use these pre-trained models can help avoid the dependence on the
complete training dataset and save the training cost. 

\begin{figure}[h]
\centering
\includegraphics[width=0.95\textwidth]{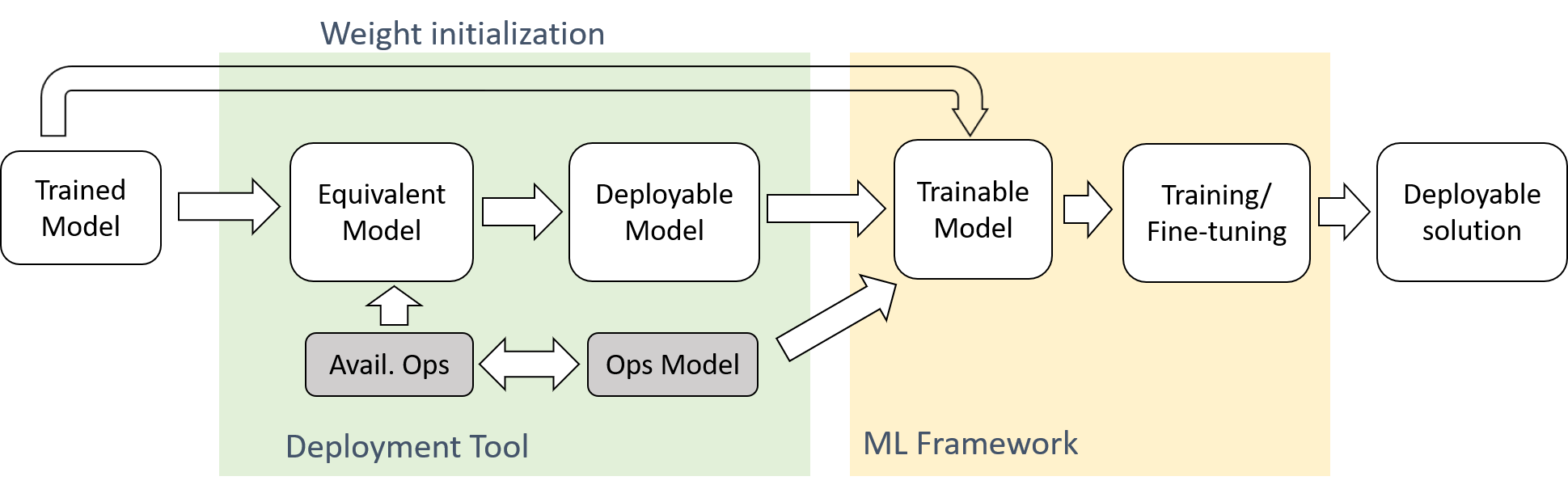}
\caption{\label{fig:retrofit}Pre-trained model deployment using proposed approach.}
\end{figure}

Our proposed approach can also be used to deploy a pre-trained model. The deployment 
flow is shown in Fig.~\ref{fig:retrofit}. The flow is similar to current approach
(shown in Fig.~\ref{fig:currappr}) where (a). operators in the trained model have to be mapped
into the implemented ones, and (b). trained weights/parameters are converted into the formats
that match the operator implementation. But the quality requirements (i.e., accuracy loss) of 
these conversions can be relaxed as it offers re-training capabilities that may recover 
the accuracy loss.

\section{Conclusion}
\label{sec:conclusion}
Big data and ample compute power have been widely considered as the two key enablers for 
deep learning. Though typically overlooked, the availability of proper tooling, such as
the ML frameworks, is also important to enable more developers to experiment new ideas
more easily and productively.

As more and more ML development and deployment targets edge devices,
there is the question that whether current ML development and deployment approach offers
enough flexibility allowed in the implementation and guarantee quality during the 
deployment. In this work, we propose a new ML development and deployment approach to
address these issues. We build a prototype KANJI for Arm Cortex-M CPUs with CMSIS-NN and 
demonstrate that this proposed approach can remove all the deployment challenges and generate 
solutions that are more efficient and with better quality.

%\bibliography{paper}
%\bibliographystyle{unsrt}

\end{document}